\documentclass[10pt,twocolumn,letterpaper]{article}

\usepackage{cvpr}
\usepackage{times}
\usepackage{epsfig}
\usepackage{graphicx}
\usepackage{amsmath}
\usepackage{amssymb}
\usepackage{caption}
\usepackage{multirow}
\usepackage{arydshln}
\usepackage[accsupp]{axessibility}  

\usepackage[pagebackref=true,breaklinks=true,letterpaper=true,colorlinks,bookmarks=false]{hyperref}

\cvprfinalcopy 

\def\cvprPaperID{****} 

\ifcvprfinal\fi
\begin{document}

\title{Generalizable Human Pose Triangulation}

\author{Kristijan Bartol\qquad David Bojani\'{c}\qquad Tomislav Petkovi\'{c}\qquad Tomislav Pribani\'{c}\\
University of Zagreb, Faculty of Electrical Engineering and Computing, Croatia\\
{\tt\small name.surname@fer.hr}

}

\maketitle
\thispagestyle{empty}

\begin{abstract}
    We address the problem of generalizability for multi-view 3D human pose estimation. The standard approach is to first detect 2D keypoints in images and then apply triangulation from multiple views. Even though the existing methods achieve remarkably accurate 3D pose estimation on public benchmarks, most of them are limited to a single spatial camera arrangement and their number. Several methods address this limitation but demonstrate significantly degraded performance on novel views. We propose a stochastic framework for human pose triangulation and demonstrate a superior generalization across different camera arrangements on two public datasets. In addition, we apply the same approach to the fundamental matrix estimation problem, showing that the proposed method can successfully apply to other computer vision problems. The stochastic framework achieves more than 8.8\% improvement on the 3D pose estimation task, compared to the state-of-the-art, and more than 30\% improvement for fundamental matrix estimation, compared to a standard algorithm.
\end{abstract}

\section{Introduction}

\noindent 
Human pose estimation is a vision task of detecting the keypoints that represent a standard set of human joints. The area is extremely competitive, especially due to the advances in deep learning. 
Pose estimation is particularly important for applications such as medicine, fashion industry, anthropometry, and entertainment \cite{a-review-of-body-measurement}. In this work, we focus on 3D human pose estimation from multiple views in a single time frame.

\begin{figure}[t!]
    \centering
    \includegraphics[width=.95\linewidth]{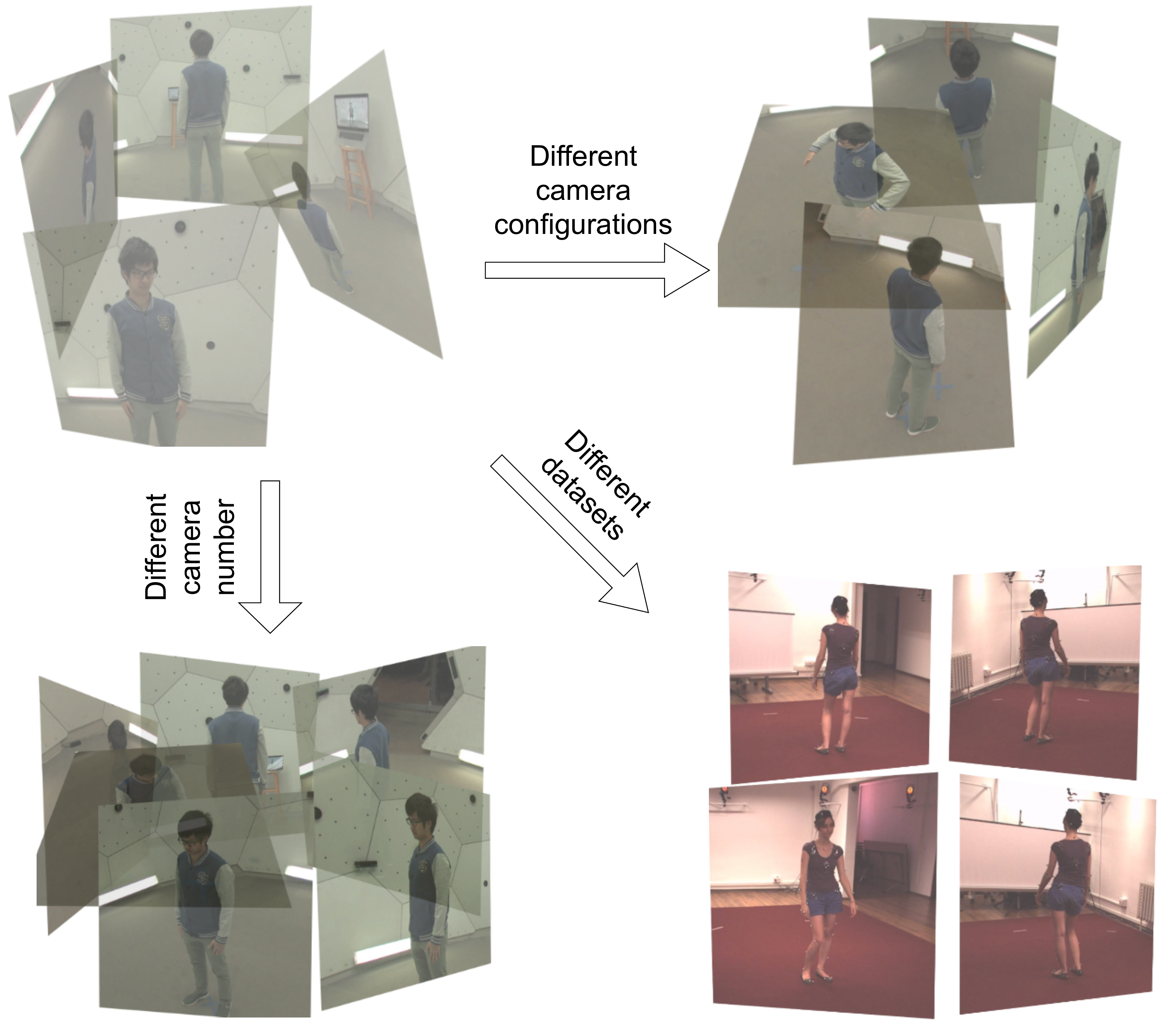}
    \caption{We propose a stochastic framework for human pose triangulation from multiple views and demonstrate its successful generalization across different camera arrangements, their number, and different public datasets. The upper two and the lower left image shows different camera arrangements and their number on CMU Panoptic Studio dataset \cite{cmu-panoptic}. The lower right part shows the Human3.6M's 4-camera arrangement \cite{h36m}.}
    \label{fig:transfer-learning}
\end{figure}

The common approach to multi-view pose estimation is to (1) detect correspondent 2D keypoints in each view using pretrained pose detector \cite{simple-baselines, openpose, cpm}, and then (2) triangulate \cite{learnable-triangulation, cross-view-fusion, epipolar-transformers, lightweight-multi-view, generalizable-approach, rethinking-pose-in-3d}. A naive approach takes 2D detections as they are and applies triangulation from all available views. 
Due to the variety of poses and self-occlusions, some views contain erroneous detections, which should be ignored or their influence mitigated in the triangulation process.
One way to ignore the erroneous detections is to apply RANSAC \cite{ransac}, marking the keypoints whose reprojection errors are above a threshold as outliers \cite{multiview-bootstrapping, epipolar-transformers}. The problem with vanilla RANSAC is that it is non-differentiable, so the gradients are not back-propagated, which disables end-to-end learning. 
Most of the state-of-the-art 3D pose estimation approaches extract 2D image features, such as heatmaps, from multiple views, and combine them for 3D elevation in an end-to-end fashion \cite{learnable-triangulation, lightweight-multi-view, cross-view-fusion}; we refer to those approaches as the \textit{learnable triangulation approaches}.

Due to a mostly-fixed set of cameras during training, the learnable triangulation approaches are often limited to a single camera arrangement and their number. Several works attempt to generalize outside the training data \cite{lightweight-multi-view, learnable-triangulation, generalizable-approach, epipolar-transformers, view-invariant-probabilistic-embedding, adaptively-multi-view-transformer, voxelpose}, but the demonstrated performance on novel views is significantly lower than using the original (base) views.

Inspired by stochastic learning \cite{stochastic-computation-graphs} and its applications in computer vision \cite{dsac, less-is-more, ng-ransac}, we propose \textit{generalizable triangulation} of human pose. First, we generate a pool of random hypotheses. A hypothesis is a 3D pose where the points are obtained by triangulating a random subset of views for each joint separately. Each generated hypothesis pass through a scoring neural network. The loss function is an expectation of the triangulation error, i.e. $\mathbb{E} (h_i) = \sum_i e_i s_i$, where $e_i$ is the error of the hypothesis $h_i$ and $s_i$ is the hypothesis score. By minimizing the error expectation, the model learns the distribution of hypotheses. The key idea is to learn to evaluate 3D pose hypotheses without considering the spatial camera arrangement used for triangulation.

The proposed approach has several practical advantages over the previous methods. First, we demonstrate its consistent generalization performance across different camera arrangement on two public datasets - Human3.6M \cite{h36m} and Panoptic Studio \cite{cmu-panoptic} (see Fig. \ref{fig:transfer-learning}).
Second, we show that the proposed model learns human pose prior and define a novel metric for pose prior evaluation. Finally, we apply the same stochastic approach to the problem of fundamental matrix estimation from noisy 2D detections and compare it to the standard $8$-point algorithm, showing that the proposed framework successfully applies to computer vision problems other than human pose triangulation.


\section{Related Work}

We distinguish two types of related work. First, we focus on triangulation-based 3D pose estimation methods and methods that attempt to generalize between the different camera arrangements and datasets. Second, we relate to keypoint correspondence methods and point out how our problem differs from the standard correspondence problem.

\textbf{Triangulation.} 
Most of the single-person image-based approaches either use robust triangulation (RANSAC) or apply learnable triangulation. Several methods \cite{self-supervised-3d-pose-using-multi-view-geometry, multiview-bootstrapping, interhand-2.6m} based on robust triangulation use RANSAC on many (more than four) views to apply triangulation only on inlier detection candidates to produce pseudo ground truth data. He et al. \cite{epipolar-transformers} exploit epipolar constraints to find the keypoint matches between multiple images and then apply robust triangulation.

The standard approach for learnable triangulation using deep learning models \cite{learnable-triangulation, cross-view-fusion, rethinking-pose-in-3d, real-time-3d-pose-smart-edge-sensors, self-supervised-with-multiple-view} is to first extract 2D pose heatmaps, where each heatmap represents the probability of a keypoint location. Cross-view fusion \cite{cross-view-fusion} builds upon the pictorial structures model \cite{3d-pictorial-structures} to combine 2D keypoint features from multiple views to estimate a 3D pose. An algebraic triangulation \cite{learnable-triangulation} estimates the confidence for each keypoint detection and applies weighted triangulation. Their volumetric approach combines the multi-view features and builds the volumetric grid, obtaining the current state-of-the-art for single-frame 3D pose. Finally, \cite{lightweight-multi-view} fuses the features into a unified latent representation that is less memory intensive than the volumetric grids. Similar to us, they also attempt to disentangle from the specific spatial camera arrangement.

\textbf{Keypoint correspondence.} The standard keypoint-based computer vision approaches, such as structure-from-motion \cite{colmap}, rely on sparse keypoint detections to establish initial 3D geometry. The core problem is to determine the correspondences between the extracted keypoint detections across images, under various illumination changes, texture-less surfaces, and repetitive structures \cite{mvs-tutorial}. The usual approach is to apply keypoint descriptor such as SIFT \cite{sift} and find inlier correspondences using RANSAC \cite{ransac}. Even though this paradigm is successful in practice, it is not differentiable and, therefore, cannot be used in an end-to-end learning fashion.

Several works have proposed soft and differentiable versions of RANSAC (DSAC) \cite{dsac, less-is-more, ng-ransac, learning-to-find-good-correspondences}. The successful soft RANSAC alternative \cite{learning-to-find-good-correspondences} learns to extract both local features of each data point, as well as retain the global information of the 3D scene. Similar to us, they also demonstrate convincing generalization capabilities to unseen 3D scenes. On the other hand, DSAC and its variants \cite{dsac, less-is-more, ng-ransac} propose a probabilistic learning scheme, i.e. minimizing the error expectation. We follow their approach but also discover that different strategies work better for our problem (see Sec. \ref{sec:method} and \ref{sec:experiments}).

In contrast to the standard keypoint matching approaches, we extract keypoints with already known human joint correspondences between the views. However, our correspondent keypoints are noisy, oscillating around the centers of the joints, which potentially leads to erroneous triangulation. Our model demonstrates robustness to erroneous keypoint detections. 

\section{Method}
\label{sec:method}

\begin{figure*}[t!]
	\centering
	\includegraphics[width=.9\linewidth]{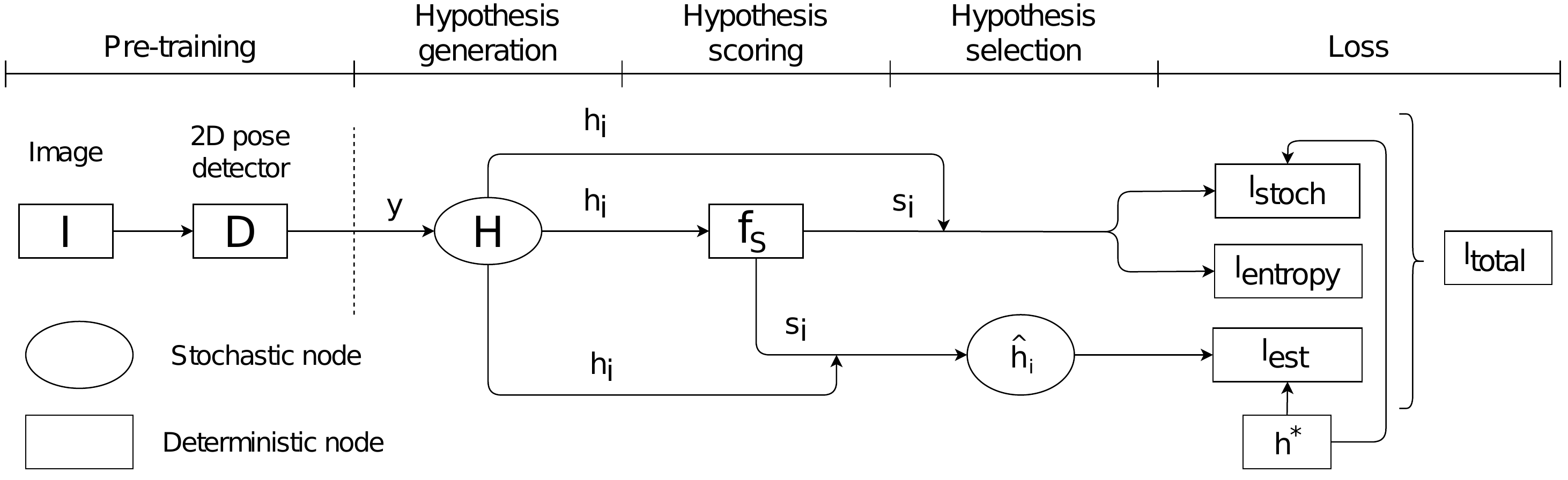}
	\caption{An overview of our method. Before stochastic learning, 2D keypoints, $y$, are extracted. In each frame, the hypothesis pool, $h_i \in \mathbf{H}$, is generated, and the poses are passed through the scoring network, $f_S$. The hypothesis $\hat{h}_i$ is selected based on the estimated scores $s_i$. Finally, the total loss, $l_{\textit{total}}$, consists of three components ($l_{\textit{stoch}}$, $l_{\textit{entropy}}$, $l_{\textit{est}}$), and is calculated with respect to the ground truth, $h^{*}$.}
	\label{fig:method}
\end{figure*}

We first describe the generic stochastic framework, and then describe it more specifically for generalizable pose triangulation and fundamental matrix estimation. The framework consists of several steps, shown in Fig. \ref{fig:method}:

\begin{enumerate}
    \item \textbf{Pre-training.} Prior to stochastic learning, the 2D poses (keypoints) are extracted for all images in the dataset. In all our experiments, we use the keypoints extracted using a baseline model \cite{simple-baselines} pretrained on Human3.6M dataset. The input to stochastic model, therefore, consists only of keypoint detections, $\mathbf{y}$. In each frame, $J$x$K$ keypoints are detected, where $J$ is the number of joints, and $K$ is the number of views.

    \item \textbf{Hypothesis generation, $\mathbf{H}$}. As it is possible to generate an extremely large number of hypotheses, only a subset of random hypotheses is created. Following \cite{stochastic-computation-graphs} and \cite{dsac}, we model the hypothesis generation step as a \textit{stochastic} node.
    
    \item \textbf{Hypothesis scoring, $\mathbf{f_{S}}$}. Each generated hypothesis $h_i \in {\mathbf{H}}$ is scored using a scoring function, $f_S (h_i | \mathbf{y}) = s_i$. The scoring function is a neural network, i.e. a multi-layer perceptron. The network architectures for 3D pose triangulation and fundamental matrix estimation differ and are specified at the end of the Sec. \ref{sec:experiments}. The network is the only learnable part of our model. The estimated scores $s_i$, passed through the Gumbel-Softmax, $\sigma_{GS}(s_i)$ (Eq. \ref{eq:gumbel-softmax}), represent the estimated probability distribution of the hypotheses $\mathbf{H}$, $\mathbb{\theta_{\mathbf{H}}}$.
    
    \item \textbf{Hypothesis selection, $\hat{h}_i$}. We experiment with several hypothesis selection strategies. The one that works the best for us is the weighted average of all hypotheses:
    
        \begin{equation}
        \label{eq:weighted}
            \hat{h}_{\textit{weight}} = \sum_i s_i h_i, \quad \sum_i s_i = 1, \quad h_i \in \mathbf{H},
        \end{equation}
    
    where the scores $s_i$ are used as weights. We also try other strategies, such as the stochastic selection:
    
    \begin{equation}
        \label{eq:stoch}
            \hat{h}_{\textit{stoch}} = h_i, \quad \text{with} \quad i \sim \theta_{\mathbf{H}},
    \end{equation}
    
    where hypothesis $h_i$ is selected based on the estimated distribution $\theta_{\mathbf{H}}$. As shown in Sec. \ref{sec:experiments}, the stochastic selection performs worse than the weighted, in contrast to \cite{dsac}.
    
    \item \textbf{Loss calculation, $l_{total}$}. The loss function consists of several components: 
    \begin{enumerate}
        \item Stochastic loss. Following \cite{dsac}, we calculate our stochastic loss as an expectation of error for all hypotheses, $l_{\textit{stoch}} = \mathbb{E} (e_{\mathbf{H}}) = \sum_i e (h_i, h^*) s_i$, where $e_i$ is the error of the estimated hypothesis with respect to the ground truth, $h^*$, and $s_i$ represent the probability that the error is minimal.
        \item Entropy loss. Score estimations $s_i$ tend to quickly converge to zero. To stabilize the estimation values, we follow \cite{less-is-more} and minimize an entropy function, $l_{\textit{entropy}} = -\sum_i s_i \log(s_i)$.
        \item Estimation loss. We define it as the error of the selected hypothesis with respect to the ground 3D pose, $l_{\textit{est}} = e_i(\hat{h}_i, h^*)$. The estimation loss, in the case of generalizable pose triangulation, is most similar to the standard 3D pose estimation loss, used by the competing approaches \cite{learnable-triangulation, lightweight-multi-view, cross-view-fusion, rethinking-pose-in-3d, generalizable-approach}.
    \end{enumerate}
    
    Finally, the total loss is a sum of the three components, $l_{total} = \alpha \, l_{\textit{stoch}} + \beta \, l_{\textit{entropy}} + \gamma \, l_{\textit{est}}$, where $\alpha$, $\beta$, and $\gamma$ are fixed hyperparameters that regulate relative values between the components.
\end{enumerate}

In order for the estimated scores $s_i$ to represent the probabilities, their values need to be normalized into $[0, 1]$ range. The standard way to normalize the output values is to apply the softmax function, $\sigma(s_i) = \frac{\exp{s_i}}{\sum_j \exp{s_j}}$. To avoid early convergence, we use the Gumbel-Softmax function \cite{gumbel-softmax, concrete-distribution}:

\begin{equation}
\label{eq:gumbel-softmax}
    \sigma_{GS}(s_i) = \frac{\exp ((\log{s_i} + g_i) / \tau)}
    {\sum_{j=1}^k \exp ((\log{s_j} + g_j) / \tau)},
\end{equation}

where $\tau$ is a temperature parameter, and $g_i$ represent samples drawn from \textit{Gumbel}(0, 1) \cite{a-star-sampling} distribution. The temperature $\tau$ regulates the broadness of the distribution. For lower temperatures ($\tau < 1$), the influence of lower-score hypotheses is limited compared to higher-score hypotheses, and vice versa. 
The purpose of \textit{Gumbel}(0, 1) is to add noise to each sample while retaining the original distribution(s), which allows the model to be more flexible with the hypothesis selection.

\subsection{Generalizable Pose Triangulation}
\label{subsec:pose-triangulation-learning}

We now describe the stochastic framework specifically for learning human pose triangulation. 

\textbf{Pose generation.} The 3D human pose hypothesis, $h_i \in \mathbf{H}$, is generated in the following way. For each joint $k$, a subset of views, $\mathbf{v}_k$, is randomly selected. The detections from the selected views are triangulated to produce a 3D joint.

\textbf{Pose normalization.} The input to the pose scoring network, $f_{S, \textit{pose}}$ are 3D pose coordinates, $\mathbf{p}$, normalized in the following way --- we select three points: left and right shoulder and the pelvis (between the hips), calculate the rotation between the normal of the plane given by the three points, and the normal of the $xy$-plane, and apply that rotation to all coordinates. Other than the 3D pose coordinates, we also extract 16 body part lengths, given by all adjacent joints, e.g. left lower arm, left upper arm, left shoulder, etc. Finally, we concatenate both normalized 3D coordinates and the body part lengths into a 1D vector and pass it through the network. The output is a scalar, $s_i$, representing the score of the hypothesis $h_i$.

\textbf{Pose estimation error.} The pose estimation error, $e_i(\hat{h}_i, h^*)$, is a mean per-joint precision error (MPJPE) \cite{h36m} between the estimated 3D pose, $\hat{\mathbf{p}}_i$, and the ground truth, $\mathbf{p}^*$:

\begin{equation}
\label{eq:mpjpe}
    e_i(\hat{h}_i, h^*) = e_i(\hat{\mathbf{p}}_i, \mathbf{p}^*) = \frac{1}{J} \sum_k^{J} ||\hat{p}_{ik} - p_{k}^*||_2,
\end{equation}

where $p_{ik}$ is the $k$-th keypoint of the $i$-th pose.

\subsection{Fundamental Matrix Estimation}
\label{subsec:learning-self-calibration}

We describe how to learn fundamental matrix estimation between the pairs of cameras using the proposed stochastic framework. The fundamental matrix describes the relationship between the two views via $x_2^{\top} F x_1 = 0$, where $x_1$ and $x_2$ are the corresponding 2D points in the first (target) and the second (reference) view. From the fundamental matrix, relative rotation and translation (the relative camera pose) between the views can be obtained \cite{zisserman}. 

\textbf{Hypothesis generation.} The relative camera pose hypothesis, $h_i$, is generated in a slightly different way than the 3D pose hypothesis. The required number of points to determine the fundamental matrix is $8$ when an $8$-point algorithm is used \cite{8-point}. However, with the presence of noise, the required number of points is usually much higher. Instead of using a single time frame as in pose triangulation, we select the keypoints from $M$ frames, having a total of $M*J$ individual point correspondences. The camera hypothesis $h_i$ is obtained using a subset of $T<M*J$ correspondences, passed through an $8$-point algorithm. The result of an $8$-point algorithm are four possible rotations and translations; we select the correct one in a standard way \cite{zisserman}.

\textbf{Input preparation.} The input to the camera pose scoring network, $f_{S, \textit{cam}}$, are the distances between the corresponding projected rays. The rays are obtained using the reference camera parameters, $(R_{ref}, t_{ref})$, and the estimated relative camera pose, $(R_{rel,i}, t_{rel,i})$. 
To achieve the permutation invariance between the line distances on the input, we simply sort the values before passing it through the network.

\textbf{Hypothesis selection.} The camera pose hypothesis, $\hat{h}_{\textit{weight}}$, is selected as the weighted average of the rotation\footnote{The rotations are converted to quaternions, for simplicity.}, i.e. a weighted average of the translation of all hypotheses. 

\textbf{Estimation error.} The hypothesis estimation error, $e_i$, is calculated as: 

\begin{equation}
\label{eq:reprojection-3d-error}
    e_i(\hat{h}_i, h^*) = e_i(\hat{\mathbf{X}}_i, \mathbf{X}^*) = ||\hat{\mathbf{X}}_i - \mathbf{X}^*||_2
\end{equation}

where $\mathbf{X}^*$ are random 3D points (used as ground truth), and $\hat{\mathbf{X}}_i$ are 3D points obtained by projecting the points $\mathbf{X}^*$ to 2D planes, using the estimated parameters, $(\hat{R}_{i}, \hat{t}_{i})$, and then projected back to 3D. More specifically, using the estimated, target projection matrix, $\hat{P}_i = K_i [\hat{R}_i | \hat{t}_i]$ and the reference projection matrix, $P_{ref} = K_{ref} [R_{ref} | t_{ref}]$, the points $\mathbf{X^*}$ are first projected to 2D, $\hat{\mathbf{x}}_i = \hat{P_i} \mathbf{X^*}$, and then triangulated using $P_{\textit{ref}}$ and $\hat{P}_i$. The intrinsic matrices $K_i$ are assumed to be known for all cameras.

\begin{table*}
\centering
\small
\captionsetup{font=small}
\caption{The demonstration of the generalization performance (MPJPE in mm) on five data sets, featuring different spatial camera placements, different number of cameras, and different datasets (CMU Panoptic Studio and Human3.6M). Each row shows the performance on five test sets when the specified train set is used. The maximal difference between the scores for particular test sets is shown in the last column. The last row demonstrates inter-dataset generalization performance, while other rows show intra-dataset performance.}
\begin{tabular}{ c | c c | c c | c c | c c | c c | c } 
\hline
Train & \multicolumn{2}{c}{CMU1} & \multicolumn{2}{c}{CMU2} & \multicolumn{2}{c}{CMU3} & \multicolumn{2}{c}{CMU4} & \multicolumn{2}{c}{H36M} & Max diff. $\downarrow$  \\
[0.5ex] 
\hline\hline
\multirow{5}{*}{Test} & CMU1 & 25.8 & CMU1 & 25.8 & CMU1 & 25.6 & CMU1 & 25.2 & CMU1 & 25.6 & 2.3\% \\

 & CMU2 & 25.4 & CMU2 & 26.0 & CMU2 & 25.5 & CMU2 & 25.6 & CMU2 & 25.9 & 2.4\% \\
 
 & CMU3 & 24.9 & CMU3 & 26.0 & CMU3 & 25.0 & CMU3 & 25.0 & CMU3 & 25.7 & 4.4\% \\
 
 & CMU4 & 25.1 & CMU4 & 25.6 & CMU4 & 25.3 & CMU4 & 25.1 & CMU4 & 25.5 & \textbf{2.0\%} \\
 
 \hdashline
 
 & \textbf{H36M} & \textbf{33.5} & H36M & 33.4 & \textbf{H36M} & \textbf{31.0} & H36M & 32.5 & \textbf{H36M} & \textbf{29.1} & 15.1\% \\
\hline\hline
\end{tabular}
\label{tab:generalization-evaluation}
\end{table*}

\begin{table}
\centering
\small
\captionsetup{font=small}
\caption{The evaluation of generalization performance from CMU Panoptic Studio \cite{cmu-panoptic} to Human3.6M dataset \cite{h36m}, compared to the volumetric approach of Iskakov et al. \cite{learnable-triangulation}. The proposed approach achieves 8.8\% better performance on H3.6M compared to \cite{learnable-triangulation}, when trained on a 4-camera CMU3 dataset (see Table \ref{tab:generalization-evaluation}).}
\begin{tabular}{ c | c | c } 
\hline
\multicolumn{3}{c}{CMU $\rightarrow$ H3.6M} \\

\hline

\textbf{Ours} & Iskakov et al. \cite{learnable-triangulation} & Improvement \\
[0.5ex] 
\hline\hline
31.0\,mm & 34.0\,mm & \textbf{8.8\%} \\

\hline\hline
\end{tabular}
\label{tab:iskakov-comparison}
\end{table}

\begin{table}
\centering
\small
\captionsetup{font=small}
\caption{The evaluation of generalization performance compared to Remelli et al. \cite{lightweight-multi-view} (lower is better). We measure the performance drop between the base test set and the novel test set for intra-dataset and inter-dataset configurations. Note that we do not compare on the same datasets, so we only measure the relative drop in percentages. Still, our approach demonstrates a significantly smaller performance drop compared to the competing method in all setups. The $\dagger$ presents the canonical fusion, and the $\ddagger$ presents the baseline approach in \cite{lightweight-multi-view}.}
\begin{tabular}{ c | c c | c } 
\hline
\multicolumn{4}{c}{Intra-dataset} \\

\hline 

Method (train dataset) & Base test & Novel test & Diff. $\downarrow$ \\

\hline\hline
Remelli et al. \cite{lightweight-multi-view} (TC1)$^\dagger$ & 27.5\,mm & 38.2\,mm & 38.9\% \\

Remelli et al. \cite{lightweight-multi-view} (TC1)$^\ddagger$ & 39.3\,mm & 48.2\,mm & 22.6\% \\

\textbf{Ours} (CMU1) & 24.9\,mm & 25.8\,mm & 3.6\% \\

\textbf{Ours} (CMU3) & 25.0\,mm & 25.6\,mm & 2.4\% \\

\textbf{Ours} (CMU4) & 25.0\,mm & 25.6\,mm & 2.4\% \\
 
\textbf{Ours} (CMU2) & 25.6\,mm & 26.0\,mm & \textbf{1.6\%} \\

\hline\hline

\multicolumn{4}{c}{Inter-dataset} \\

\hline 

Method (train dataset) & H36M & CMU1 & Diff. $\downarrow$ \\


\hline\hline

\textbf{Ours} (H36M) & 29.1\,mm & 33.5\,mm & \textbf{15.1\%} \\

\end{tabular}
\label{tab:remelli-comparison}
\end{table}

\section{Experiments}
\label{sec:experiments}

The stochastic framework is evaluated on Human3.6M \cite{h36m} and Panoptic Studio \cite{cmu-panoptic} datasets. As most of the previous 3D pose estimation approaches presented their results on Human3.6M, we use it for the quantitative comparison to state-of-the-art. Panoptic Studio contains a relatively large number of cameras (31) with useful data annotations (camera parameters, 3D and 2D poses). We use the Panoptic Studio dataset to evaluate the generalization performance between different camera arrangements and their number. We also evaluate the generalization between the Panoptic Studio and Human3.6M datasets. As experiments are based on a single-person pose estimation, we use Panoptic Studio sequences that contain single person in the scene, following \cite{monocular-total-capture}.

\begin{table}[t!]
\centering
\small
\captionsetup{font=small}
\caption{The comparison to RANSAC, algebraic triangulation \cite{learnable-triangulation}, and VoxelPose \cite{voxelpose} on Panoptic Studio (intra-dataset) [mm]. The numbers show the performance on novel camera views. Our number is obtained as an average over 12 non-diagonal values of Table \ref{tab:generalization-evaluation}.} 
\begin{tabular}{ c c c c } 
\hline
\multicolumn{4}{c}{Intra-dataset (CMU Panoptic Studio)} \\

\hline 

RANSAC & Algebraic & VoxelPose & Ours \\

\hline

39.5 & 33.4 & 25.5 & \textbf{25.4} \\

\hline\hline

\end{tabular}
\label{tab:sota-comparison}
\end{table}

Other than the evaluation of our best result ($\hat{h}_{\textit{weight}}$), we also compare between different hypotheses:

\begin{itemize}
    \item Weighted average hypothesis, $\hat{h}_{\textit{weight}}$,
    
    \item Average hypothesis,         $\hat{h}_{\textit{avg}}$, obtained as an average of all hypotheses,
    
    \item Most and least probable hypotheses, $\hat{h}_{\textit{most}}$ and $\hat{h}_{\textit{least}}$, the hypotheses with maximal and minimal estimated score, $s_{\textit{max}}$ and $s_{\textit{min}}$,
    
    \item Stochastic hypothesis, $\hat{h}_{\textit{stoch}}$, selected randomly, based on the estimated distribution $\theta_{\mathbf{H}}$,
    
    \item Random hypothesis, $\hat{h}_{\textit{random}}$, selected randomly from an uniform distribution,
    
    \item Best and worst hypotheses\footnote{Note that the best and the worst hypotheses are not available in inference (missing \string^ sign), because they are determined using ground truth.}, $h_{\textit{best}}$ and $h_{\textit{worst}}$, with the lowest and the highest errors, $e_{\textit{min}}$ and $e_{\textit{max}}$.
\end{itemize}

Additionally, we also compare ourselves with RANSAC as reported in \cite{learnable-triangulation} (see Subsec. \ref{subsec:quantitative-evaluation}).

\subsection{Generalization Performance}
\label{subsec:transfer-learning}

One of the most important properties of the proposed model is that it generalizes well to different spatial arrangements and number of cameras, and different datasets, which is a major limitation of the previous models. To evaluate the generalization performance across data sets, we select five different camera arrangements:

\begin{enumerate}
    \item Cameras $1, 2, 3, 4, 6, 7, 10$ (CMU1),
    \item Cameras $12, 16, 18, 19, 22, 23, 30$ (CMU2),
    \item Cameras $10, 12, 16, 18$ (CMU3),
    \item Cameras $6, 7, 10, 12, 16, 18, 19, 22, 23, 30$ (CMU4), and
    \item Cameras $0, 1, 2, 3$ (H36M).
\end{enumerate}

\begin{table*}[t!]
\tiny
\centering
\captionsetup{font=small}
\caption{No additional training data setup. Overall comparison to the state-of-the-art on Human3.6M dataset. The proposed method outperforms most of the state-of-the-art methods. All values are showing MPJPE scores (mm).}
 \begin{tabular}{l | c c c c c c c c c c c c c c c | c} 
 \hline
 Protocol 1, abs. positions & Dir. & Disc. & Eat & Greet & Phone & Photo & Pose & Purch. & Sit & SitD. & Smoke & Wait & WalkD. & Walk & WalkT. & Avg $\downarrow$ \\ [0.5ex] 
 \hline\hline
 Tome et al. \cite{rethinking-pose-in-3d} & 43.3 & 49.6 & 42.0 & 48.8 & 51.1 & 64.3 & 40.3 & 43.3 & 66.0 & 95.2 & 50.2 & 52.2 & 51.1 & 43.9 & 45.3 & 52.8 \\
 Kadkhodamohammadi et al. \cite{generalizable-approach} & 39.4 & 46.9 & 41.0 & 42.7 & 53.6 & 54.8 & 41.4 & 50.0 & 59.9 & 78.8 & 49.8 & 46.2 & 51.1 & 40.5 & 41.0 & 49.1 \\
 Cross-view fusion \cite{cross-view-fusion} & 28.9 & 32.5 & 26.6 & 28.1 & 28.3 & 29.3 & 28.0 & 36.8 & 41.0 & 30.5 & 35.6 & 30.0 & 28.3 & 30.0 & 30.5 & 31.2 \\
 
 Remelli et al. \cite{lightweight-multi-view} & 27.3 & 32.1 & 25.0 & 26.5 & 29.3 & 35.4 & 28.8 & 31.6 & 36.4 & 31.7 & 31.2 & 29.9 & 26.9 & 33.7 & 30.4 & 30.2 \\
 
 Epipolar transformers \cite{epipolar-transformers} & 25.7 & 27.7 & 23.7 & 24.8 & 26.9 & 31.4 & 24.9 & 26.5 & 28.8 & 31.7 & 28.2 & 26.4 & 23.6 & 28.3 & 23.5 & \textbf{26.9} \\
 
 \hline
 \textbf{Ours} ($h_{\textit{weight}}$) & 27.5 & 28.4 & 29.3 & 27.5 & 30.1 & 28.1 & 27.9 & 30.8 & 32.9 & 32.5 & 30.8 & 29.4 & 28.5 & 30.5 & 30.1 & 29.1 \\ [0.5ex] 
 
 \hline \hline
 \end{tabular}
\label{tab:quantitative}
\end{table*}

The setup is as follows. Each of the five camera arrangements is first used for training, and then the generalization performance is tested on the remaining four arrangements. 
The five selected sets differ with respect to the spatial camera arrangement and their number. Additionally, the fifth camera set (H36M) is used to test the transfer learning capabilities between the datasets. All the results in this subsection are obtained using $h_{\textit{weight}}$ hypothesis.

\textbf{Our Generalization Performance.} Table \ref{tab:generalization-evaluation} shows consistent performance on each of the five test datasets, regardless of the selected training dataset. In particular, the performance between different test sets on the Panoptic Studio dataset is within 5\% difference, which demonstrates robustness to various camera arrangements and their number (intra-dataset). The inter-dataset generalization is also successful, which we further evaluate against the competing methods \cite{learnable-triangulation, lightweight-multi-view}. Note that the demonstrated generalization can be exploited both in training time and in inference. 

\textbf{Volumetric Triangulation.} Table \ref{tab:iskakov-comparison} compares our proposed method to the state-of-the-art 3D pose estimation approach \cite{learnable-triangulation}. Iskakov et al. reported an average 34.0\,mm error on Human3.6M test set when they trained on CMU Panoptic Studio (4-camera arrangement). Compared to them, we achieve 31.0\,mm on our 4-camera arrangement (CMU3), demonstrating an improvement in inter-dataset generalization (see Table \ref{tab:generalization-evaluation} for the comprehensive results).

\textbf{Remelli et al.} Table \ref{tab:remelli-comparison} compares our method to Remelli et al. \cite{lightweight-multi-view}. Similar to us, they explicitly address the generalization to novel views. They demonstrate their intra-dataset generalization performance on Total Capture \cite{total-capture}, by comparing the test performances on cameras (1, 3, 5, 7) as a base arrangement (TC1) and testing it on cameras (2, 4, 6, 8) as a novel arrangement (TC2). We do not evaluate our model on Total Capture. Instead, to compare with Remelli et al., we measure the performance between the CMU camera test sets and evaluate relative score differences. The performance of our model is consistent across different camera arrangements and their number for intra-dataset configuration. Moreover, our inter-dataset performance from CMU Panoptic Studio to Human3.6M is 15.1\%, which is still better than the best result by Remelli et al. Note that the inter-dataset experiment is the most difficult as it also includes the changes in camera arrangement.

\textbf{RANSAC.} We outperform RANSAC on Panoptic Studio by a large margin. We can explain this by the fact that CMU does not have a full view of a person in most cameras, leading to
strong occlusions and missing parts. As RANSAC takes
only reprojection errors of individual 3D joints as an inlier
selection criterion, it is unable to evaluate the estimated 3D
pose as a whole, in contrast to our model that learns human
pose prior (see Sec. \ref{subsec:learning-human-pose-prior}).

\textbf{Algebraic Triangulation.} The algebraic triangulation \cite{learnable-triangulation} is originally proposed as an improvement over RANSAC,
where the weight is estimated for each joint location. The
weight-based model indeed outperforms RANSAC both on
Human3.6M and Panoptic Studio. However, as the authors point out in \cite{learnable-triangulation}, it has several drawbacks.
First, it processes each view independently, and second, it
separately triangulates each joint. Therefore, the weight-based algebraic model suffers from the same problem as
RANSAC by not taking the whole pose into account. Our model, on the other hand, successfully learns human pose prior, which allows it to select more feasible
poses, making it more robust to occlusions and missing
body parts. Note that \cite{learnable-triangulation} does not test their weighted model on unseen views. Therefore, Table 1 shows the result of the
model w/o weights, as this model is consistent across different camera sets. The actual result of the weighted model
might differ, but it is hard to estimate by how much.

\textbf{VoxelPose.}  VoxelPose \cite{voxelpose} reports
25.51mm MPJPE score on their intra-dataset experiment, compared to our
25.42mm. Even though we achieve
comparable performances, the significant difference is that we did not pretrain our 2D backbone on the Panoptic Studio dataset, which would most likely further improve our
2D keypoint estimation and, consequently, final 3D pose estimations (Supp.).


\subsection{Base Dataset Performance}
\label{subsec:quantitative-evaluation}

The comparison to state-of-the-art is shown in Table \ref{tab:quantitative}. Note that the Table only shows the methods that use Human3.6M for training and testing, with no additional training data (therefore, excluding Iskakov et al. \cite{learnable-triangulation}). Compared to the best-performing single-frame method, Epipolar Transformers \cite{epipolar-transformers}, we obtain $2.2$\,mm worse MPJPE, but outperform most of the other recent methods.

\begin{table}
\centering
\small
\captionsetup{font=small}
\caption{Overall quantitative comparison between the hypotheses. The values are showing MPJPE scores in mm (the lower is better).}
\begin{tabular}{ l | l l } 
\hline
Hypothesis & Human3.6M $\downarrow$ & Panoptic Studio $\downarrow$  \\
[0.5ex] 
\hline\hline
$\hat{h}_{\textit{weight}}$ & 29.1 & \textbf{24.9} \\ 
$\hat{h}_{\textit{avg}}$ & 31.2 \textcolor{green}{+2.1} & 25.9 \textcolor{green}{+1.0} \\
$\hat{h}_{\textit{most}}$ & 41.3 \textcolor{green}{+12.2} & 25.0 \textcolor{green}{+0.1} \\
$\hat{h}_{\textit{least}}$ & 74.5 \textcolor{green}{+45.4} & 29.8 \textcolor{green}{+3.9} \\
$\hat{h}_{\textit{stoch}}$ & 41.3 \textcolor{green}{+12.2} & 26.5 \textcolor{green}{+1.6} \\
$\hat{h}_{\textit{random}}$ & 45.0 \textcolor{green}{+15.9} & 26.1 \textcolor{green}{+1.2} \\
\hline
$h_{\textit{best}}$ & 22.3 -6.8 & 24.4 -0.5 \\
$h_{\textit{worst}}$ & 98.9 +69.8 & 31.0 +6.1 \\
\hline
RANSAC & \textbf{27.4 \textcolor{red}{-1.7}} & 39.5 \textcolor{green}{+14.6}\\
[0.5ex] 
\hline\hline
\end{tabular}
\label{tab:quantitative-hypotheses}
\end{table}

\begin{figure*}[t!]
    \centering
    \captionsetup{font=small}
    \includegraphics[width=.8\linewidth]{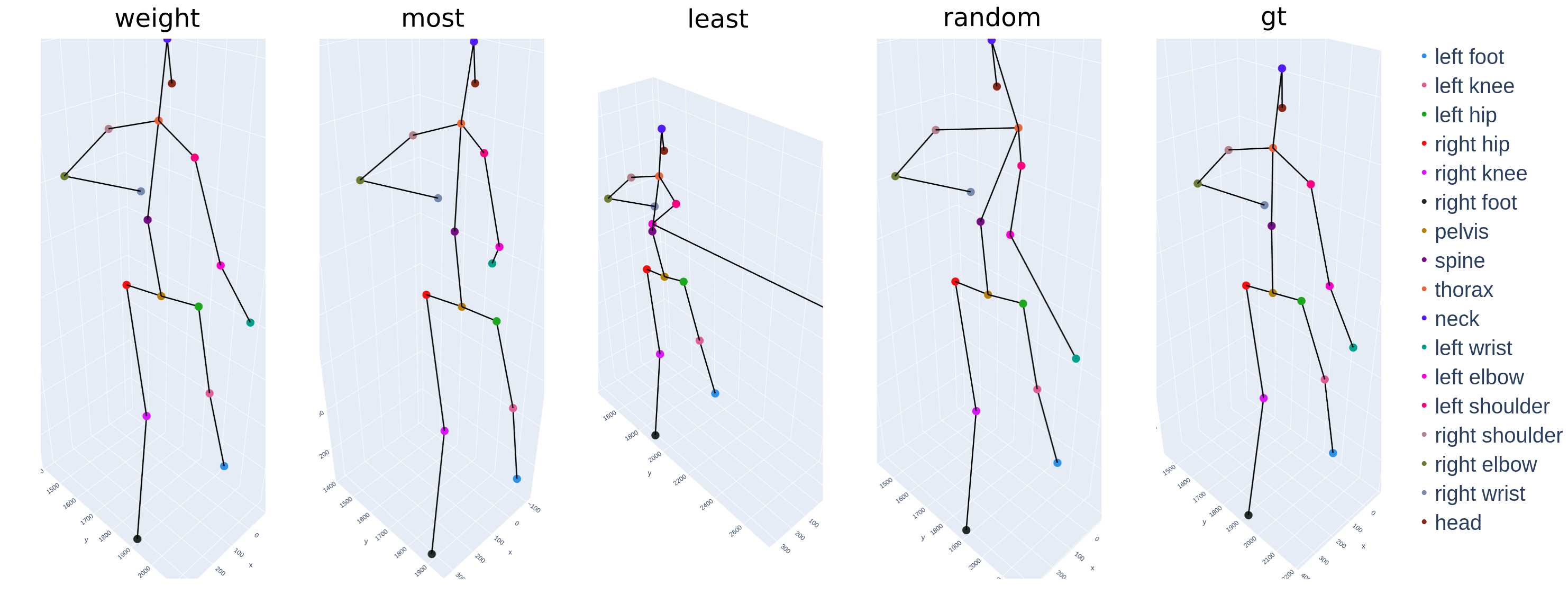}
    \caption{Qualitative comparison between four 3D pose hypotheses compared to ground truth (\textit{gt}), on Human3.6M.}
    \label{fig:qualitative}
\end{figure*}

Table \ref{tab:quantitative-hypotheses} shows the MPJPE scores of all previously described pose hypotheses on the two datasets, compared to the RANSAC result, as reported in \cite{learnable-triangulation}. Even though our weighted average hypothesis, $\hat{h}_{\textit{weight}}$, is outperformed by the RANSAC approach on Human3.6M, we show a significant improvement on Panoptic Studio. Also, note that RANSAC is competitive against most of the state-of-the-art approaches on Human3.6M that do not use additional training data. 

Regarding other results, the average hypothesis, $\hat{h}_{\textit{avg}}$ performs better than the stochastic, $\hat{h}_{stoch}$. The stochastic performs even worse than the random hypothesis on Panoptic Studio. The most probable hypothesis, $\hat{h}_{\textit{most}}$, outperforms the average on Panoptic Studio. Note that the difference between best and worst hypothesis ($h_{\textit{best}}$, $h_{\textit{worst}}$) is significantly different on the two datasets. This suggests that the hypotheses generated on Panoptic Studio are more similar to each other and the distribution is less broad. The difference between the most and the least probable hypotheses ($\hat{h}_{\textit{most}}$, $\hat{h}_{\textit{least}}$) is reasonable on both datasets, which confirms that our model learned to differentiate between the poses.

Fig. \ref{fig:qualitative} shows the qualitative performance comparison between several hypotheses. The least probable hypothesis, $h_{\textit{least}}$, does not have visually plausible pose reconstruction, while the random hypothesis, $\hat{h}_{\textit{random}}$, has some obvious errors in the upper body. The most probable hypothesis, $\hat{h}_{\textit{most}}$ has minor reconstruction errors on the right arm and shoulder. The weighted hypothesis, $\hat{h}_{\textit{weight}}$, is visually comparable to ground truth.

\begin{figure}[t!]
    \centering
    \captionsetup{font=small}
    \includegraphics[width=1.\linewidth]{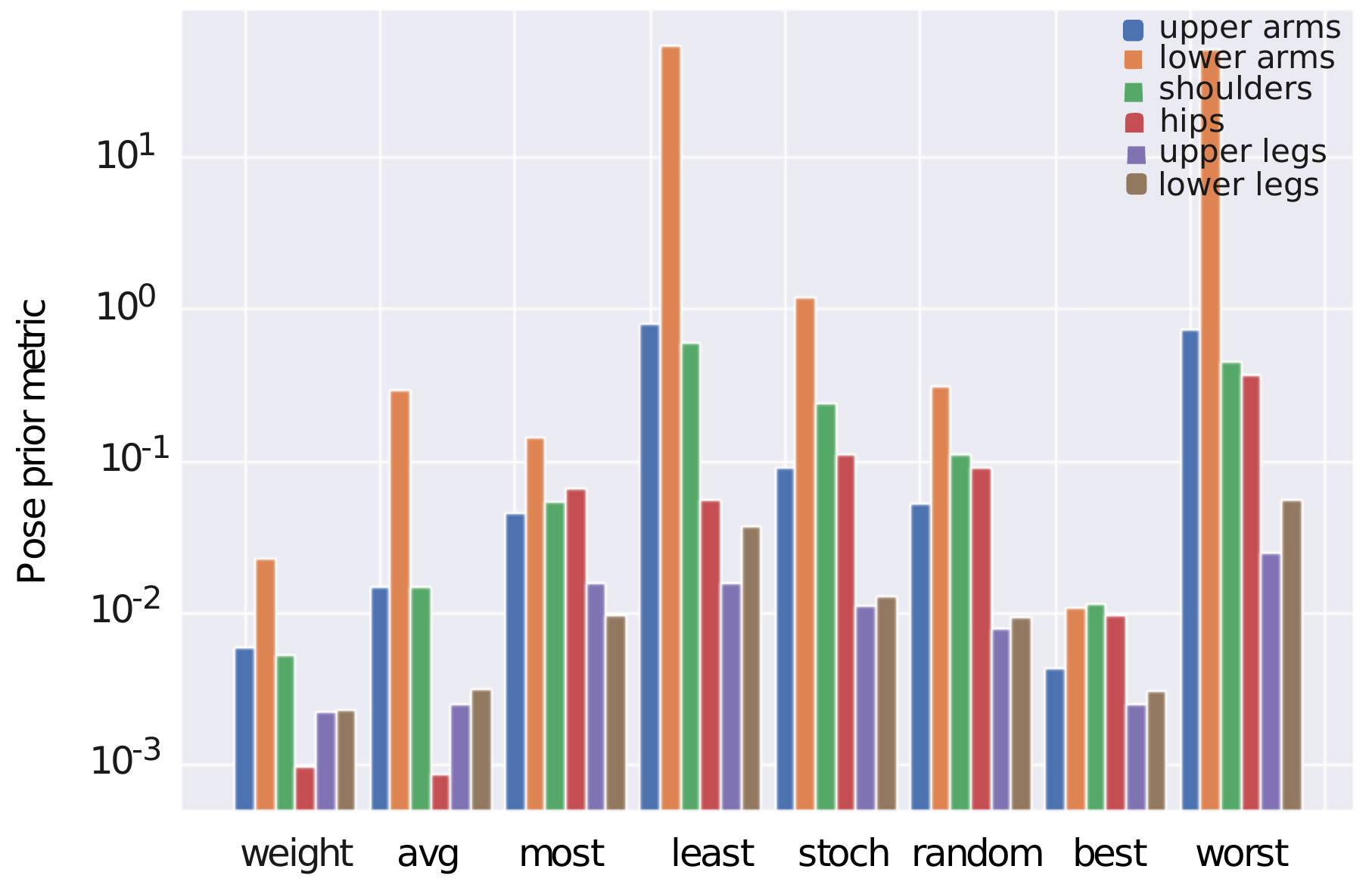}
    \caption{Evaluation of the human pose prior metric for different hypotheses, and six body part pairs (lower is better).}
    \label{fig:pose-prior}
\end{figure}

\subsection{Human Pose Prior}
\label{subsec:learning-human-pose-prior}

We demonstrate the successful pose prior learning of the pose scoring network, $f_{S, \textit{pose}}$. There are previous works that attempt learning human pose prior \cite{can-3d-pose-be-learned-from-2d, deep-kinematics-analysis, 3d-pictorial-structures}, but they do not quantitatively evaluate their methods. The idea of learning pose prior is to differentiate between the 3D poses that are more plausible and the poses that are less plausible with respect to several human body properties. The properties that can be extracted from the 3D pose are based on the body part lengths and between-joint angles. In this work, we focus on body part lengths, i.e. \textit{left-right body symmetry}.

The body symmetry is measured for six different body left-right part pairs: upper arms, lower arms, shoulders, hips, upper legs, and lower legs. For each pair, $l$, we calculate the ratio, $r_{il}$ between the left and right part, in each time frame, $i$. The final pose prior metric is a variance of the ratios over time:

\begin{equation}
    S^2 = \frac{\sum_i (r_{il} - \overline{r}_l)^2}{T-1},
\end{equation}

where $\overline{r}_l$ is the mean ratio for the pair $l$, and $T$ is the number of frames. The reason for using ratios instead of the differences between the body parts is that some people are naturally asymmetric, so the idea is only to measure the consistency over time.

Fig. \ref{fig:pose-prior} shows the pose prior metrics for the subject 9 of the Human3.6M dataset, for different hypotheses. As expected, the values are generally the lowest for our best performing hypothesis, $\hat{h}_{\textit{weight}}$, followed by the average hypothesis, $\hat{h}_{\textit{avg}}$. The difference between the most probable and the least probable hypothesis ($\hat{h}_{\textit{most}}$, $\hat{h}_{\textit{least}}$) suggests that we successfully learned body pose prior, i.e. differentiate between the plausible poses with respect to the body symmetry consistency over time. Note that the best hypothesis, $h_{\textit{best}}$, is comparable to $\hat{h}_{\textit{weight}}$.

\subsection{Fundamental Matrix Estimation}

Table \ref{tab:fundamental-evaluation} shows the fundamental matrix estimation results on all 4-view combinations on Human3.6M. 
The four metrics are used for the evaluation:

\begin{itemize}
    \item Rotation error, $E_R = ||\text{quat}(\hat{R}_{\textit{rel}}) - \text{quat}(R_{\textit{rel}}^*)||_2$, where quat() represents the conversion to quaternions,
    
    \item Translation error (mm), $E_t = ||\hat{t}_{\textit{rel}} - t_{\textit{rel}}^*||_2$,
    
    \item 2D error (pixels), $E_{2D} = ||\hat{\mathbf{x}}_i - \mathbf{x}^*||_2$, where $\mathbf{x^*}$ represents random 3D points, $\mathbf{X}^*$, projected using the ground truth relative projection matrix, $P^*$, and
    
    \item 3D error (mm), $E_{3D} = e_i (\hat{h}_i, h^*)$, from Eq. \ref{eq:reprojection-3d-error}.
\end{itemize}

The obtained results show that the model achieves subpixel error ($E_{2D}$) for two pairs of views ((1, 2) and (2, 3)), and only few pixels in the worst case, which corresponds to several millimeters when reprojected back to 3D ($E_{3D}$). Note that the adjacent pairs of views have lower errors than the opposite pairs, as expected.

\begin{table}
\centering
\small
\captionsetup{font=small}
\caption{Evaluation of fundamental matrix estimation for all pairs of views on Human3.6M, based on four error metrics. Note that the camera pairs (1, 3), and (2, 4) are diagonal, while other pairs are adjacent.}
\begin{tabular}{ c | c c c c } 
\hline
 Camera pair & $E_R \downarrow$ & $E_t \downarrow$ & $E_{2D} \downarrow$ & $E_{3D} \downarrow$  \\
[0.5ex] 
\hline\hline

(1, 3) & 1.7e-2 & 2.3e+1 & 3.7e+0 & 2.3e+1 \\

(2, 4) & 3.2e-2 & 4.9e+0 & 1.3e+0 & 2.3e+1 \\

(1, 4) & 9.8e-3 & 2.7e+1 & 2.3e+0 & 1.3e+1 \\

(2, 3) & \textbf{2.1e-3} & 9.9e+0 & \textbf{7.2e-1} & 1.3e+1 \\

(3, 4) & 4.7e-3 & 8.5e+0 & 1.2e+0 & 5.5e+0 \\

(1, 2) & 4.8e-3 & \textbf{4.8e+0} & 8.1e-1 & \textbf{4.9e+0} \\ 
[0.5ex] 
\hline\hline
\end{tabular}
\label{tab:fundamental-evaluation}
\end{table}

\begin{figure}[t!]
    \centering
    \captionsetup{font=small}
    \includegraphics[width=1.\linewidth]{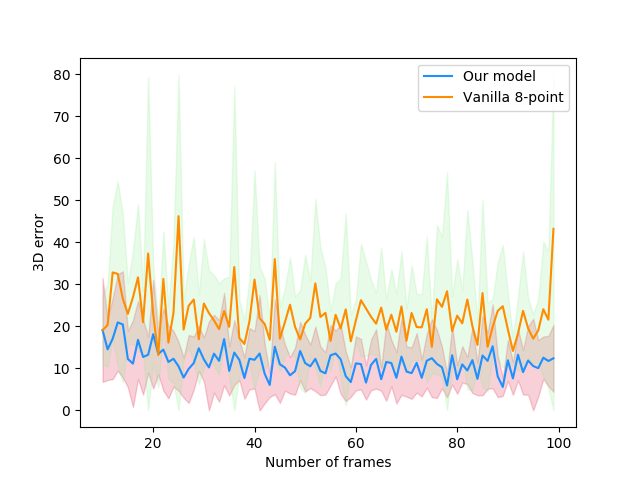}
    \caption{The comparison of the $E_{3D}$ errors between the stochastic model and the $8$-point algorithm, for different number of input frames (between $10$ and $100$), using the camera pair $(2, 3)$ of Human3.6M. For every number of frames, the experiments is done 10 times. The lines show mean values, and the fill parts show standard deviations. The values are clipped to 80\,mm.}
    \label{fig:comparison-to-8-point}
\end{figure}

In Fig. \ref{fig:comparison-to-8-point}, we compare our 3D errors ($E_{3D}$) to the vanilla $8$-point algorithm, on the (2, 3) camera pair, using different number of input frames. Our model consistently outperforms the $8$-point algorithm, showing robustness to noise and increased confidence due to lower variance.

\section{Conclusion}

The proposed generalizable approach is a promising novel direction for 3D human pose estimation, as well as other related computer vision problems, such as the camera pose estimation. 
The demonstrated results show convincing generalization capabilities between different camera arrangements and datasets, outperforming previous methods. The model requires relatively little training data, which makes training faster and more convenient for smaller datasets, as further discussed in Supplementary.

The overall performance is competitive in both human pose triangulation and camera pose estimation tasks. By combining these two steps, it is possible to transfer the performance of the base dataset to any novel multi-camera dataset, in inference. The next reasonable step is to exploit image features in an end-to-end learning fashion, which should further improve the performance and possibly outperform the state-of-the-art even on the base dataset. 
The current model supports only a single-person pose triangulation. To extend to multi-person, we need to solve the keypoint correspondence problem between the people.

\section*{Acknowledgement}

This work has been supported by the Croatian Science Foundation under the project IP-2018-01-8118.


\section*{Supplementary Appendix}
\label{appendix}

The main focus of the Supplementary Appendix is to demonstrate the application of the proposed model to novel camera arrangements and datasets that have unknown relative camera poses, i.e. extrinsic parameters $E_{\textit{ref, i}}=[R_{\textit{ref, i}}|t_{\textit{ref, i}}]$, where \textit{ref} is the reference camera, and $i$ is each of the relative cameras. The camera poses are estimated based on the fundamental matrix estimation method described in the main paper. We further dissect relative camera pose estimation into the estimation of relative rotation, $R_{\textit{ref, i}}$, and relative translation, $t_{\textit{ref, i}}$, showing that the unknown translations have more significant impact on the performance than the unknown rotations (Appendix A). 
Finally, we briefly discuss other works, implementation details, and the limitations of the model in more detail and propose future work, in addition to the main paper (Appendix D). The ethical considerations are addressed in Appendix E.

\section*{A. Performance with Estimated Camera Poses}

We evaluate the performance of the generalizable human pose triangulation model in case when the camera poses are estimated using the proposed fundamental matrix estimation method, on Human3.6M. In particular, we compare the performances between the test sets with known extrinsics, estimated relative rotations $R_{\textit{ref, i}}$, estimated relative translations, $t_{\textit{ref, i}}$, and estimated extrinsics (both rotation and translation). Additionally, we compare the performances when Human3.6M is used as the training dataset (base-dataset experiment), and when CMU3, described in the main paper, is used for training (inter-dataset experiment). 

The results are shown in Table \ref{tab:unknown-extrinsics-evaluation}. As expected, the performance on the base dataset is better than the performance on inter-dataset experiment. In overall, the performances on both the base experiment and inter-dataset experiment are satisfactory, taking into account that the rotations, translations, i.e. both, are unknown. Notably, the performance of the model significantly drops for unknown relative translations, while the unknown relative rotations only slightly affect the performance. We assume that the rotations are simply estimated more accurately than translations, hence the difference. To verify this assumption, we analyze 2D and 3D errors, defined in the main paper, for estimated rotations, i.e., translations separately.

\begin{table}[t!]
\centering
\small
\caption{The evaluation of the model in case of unknown relative camera poses on Human3.6M \cite{h36m}. We evaluate the model in base-dataset (same camera arrangement for training and testing) and inter-dataset (from CMU3 \cite{cmu-panoptic} to Human3.6M). We also dissect the analysis into the cases when rotation, i.e., translation only is unknown. Note that all $R$s and $t$s shown in the table correspond to $R_{\textit{ref, i}}$ and $r_{\textit{ref, i}}$, but are abbreviated.} 
\begin{tabular}{ c c c c } 
\hline
\multicolumn{4}{c}{Base dataset (Human3.6M)} \\

\hline 

Known $[R|t]$ & Estimated $R$ & Estimated $t$ & Estimated $[R|t]$ \\

\hline

\textbf{29.1\,mm} & 29.4\,mm & 36.7\,mm & 37.3\,mm \\

\hline\hline

\multicolumn{4}{c}{Inter-dataset (CMU3 $\rightarrow$ Human3.6M)} \\

\hline

Known $[R|t]$ & Estimated $R$ & Estimated $t$ & Estimated $[R|t]$ \\

\hline

\textbf{31.0\,mm} & 33.6\,mm & 42.2\,mm & 44.5\,mm

\end{tabular}
\label{tab:unknown-extrinsics-evaluation}
\end{table}

\textbf{Ablative Analysis of Camera Pose Estimation.} Table \ref{tab:fundamental-dissection} shows the fundamental matrix estimation errors ($E_{2D}$ and $E_{3D}$, described in the main paper) between the pairs of views, in case when only rotation is estimated and the translation is known, and vice versa. The errors in case of the estimated translations are always higher compared to the case of estimated rotations, therefore, this result might explain the performance drop shown in Table \ref{tab:unknown-extrinsics-evaluation}. The future work should focus on improving translation estimation.

\begin{table}[h!]
\centering
\small
\caption{Dissecting the evaluation of fundamental matrix estimation on two cases --- when the rotations, i.e., the translations are estimated, for all pairs of views on Human3.6M. The 2D errors, $E_{2D}$ are shown in pixels, and 3D errors, $E_{3D}$ are shown in millimeters.}
\begin{tabular}{ c | c c || c c } 

\hline
& \multicolumn{2}{c}{Estimated $R$} & \multicolumn{2}{c}{Estimated $t$} \\

\hline
 Camera pair & $E_{2D}$ & $E_{3D} $ & $E_{2D}$ & $E_{3D}$  \\
[0.5ex] 
\hline\hline

(1, 3) & 1.2 & 10.8 & 1.8 & 18.2 \\

(2, 4) & 0.9 & 9.7 & 1.6 & 15.3 \\

(1, 4) & 0.9 & 6.4 & 1.2 & 8.9 \\

(2, 3) & 0.6 & 3.9 & 1.0 & 4.5 \\

(3, 4) & \textbf{0.4} & \textbf{1.2} & \textbf{0.7} & 4.0 \\

(1, 2) & \textbf{0.4} & 1.8 & \textbf{0.7} & \textbf{3.7} \\ 
[0.5ex] 
\hline\hline
\end{tabular}
\label{tab:fundamental-dissection}
\end{table}


\section*{B. Other Works}

The Epipolar Transformers \cite{epipolar-transformers} outperforms our method on Human3.6M (base dataset). However, note that our model outperforms their lightweight, transformer model on H36M (30.4mm, Table 6 \cite{epipolar-transformers}, compared to our 29.1mm, Table 4, main paper). The difference in performances would most likely increase when evaluated on novel views, especially as the authors did not tackle the generalization problem at all. Further, their heavy-weight model might overfit even more on the base camera arrangement of the train dataset(s), so we can expect an increased performance drop on unseen views.

\section*{C. Implementation Details}

The selected hyperparameters set is shown in Tab. \ref{tab:hyperparameters}. The two hyperparameters used specifically for pose triangulation, i.e., fundamental matrix estimation, are the number of joints in the pose model, $J=17$, and the number of frames from which the camera hypotheses are sampled, $M=80$.

The required number of training iterations is relatively small. We obtain our best results using only $500$ iterations. In each iteration, we generate $200$ hypotheses. This is a great advantage of the approach, especially when only small amount data annotations are required. In particular, $500$ iterations correspond to $500$ data samples, i.e., 500/16$\approx$32 batches (batch size 16, Table \ref{tab:hyperparameters}), meaning that the gradients were applied $\approx$32 times for the model to be fully trained. It takes about 3 minutes to train the model, but this can be further improved by more efficient implementation of the hypothesis generation on CPU. Moreover, the training time is shorter, which simplifies the optimal hyperparameter search. Finally, the current implementation fits into $\sim$1GB of GPU memory.

\begin{table}
\centering
\small
\captionsetup{font=small}
\caption{The Table of hyperparameters for the two tasks.}
\begin{tabular}{ c | c c } 
\hline
 & 3D pose & Camera \\
[0.5ex] 
\hline\hline
Learning rate & $5*10e^{-4}$ & $10e^{-5}$ \\ 
$\tau$ & $1.5$ & $1.2$ \\ 
$\alpha$, $\beta$, $\gamma$ & ($1.0$, $0.01$, $0.02$) & ($1.0$, $0.01$, $0.0$) \\
\multirow{ 2}{*}{Network layer sizes} & ($1000$, $900$ & ($1000$,  \\
 & $900$, $900$, $700$) & $900$, $900$) \\
\# hypotheses in sample & 200 & 100 \\
Batch size & 16 & 16 \\
[0.5ex] 
\hline\hline
\end{tabular}
\label{tab:hyperparameters}
\end{table}

\section*{D. Limitations}
\label{appendix-3}

The main limitation of our model is that it strongly depends on the performance of the 2D detector \cite{simple-baselines}. This is best seen in Table \ref{tab:limitation-table} that shows the difference in the performance on train, validation, and test\footnote{Note that, for training, we use subjects 1, 5, 6, 7, for validation, we use subject 8, and the remaining subjects 9 and 11 are used for testing.}. The difference between the validation and test performance, in particular, can be explained by the fact that the 2D backbone has been fine-tuned on the whole training and validation splits, while it has never seen the test data. What this means is that we did not tackle the problem of train-to-test generalization; instead, we improved the between-test-sets generalization, which is a weaker result. The consequence of this train-test difference is that the performance on novel data will suffer mostly from the performance drop of the detector.

\begin{table}
\centering
\small
\caption{The comparison between train, validation, and test performance on Human3.6M (in case of base-dataset configuration). There is a significant difference in the performance between train (validation) and test.}
\begin{tabular}{ c c | c } 

\hline
\multicolumn{3}{c}{Human3.6M} \\
\hline

Train & Validation & Test \\
[0.5ex] 
\hline\hline
\textbf{13.8\,mm} & 14.2\,mm & 29.1\,mm \\

\hline\hline
\end{tabular}
\label{tab:limitation-table}
\end{table}

Another limitation is that the current model does not learn end-to-end. The consequence is that the model, at best, learns to differentiate well between the poses. But once the poses are good enough, the network can't differentiate further and will simply assign the same scores, converging into an average of "good-enough" 3D poses\footnote{The good poses should be the ones that are symmetric and the ones that have body part ratios consistent with the ratios of an average (training set) person. Note that the good poses should have high pose prior scores.}. Therefore, future work should definitely address this limitation by exploiting image features to obtain additional information about the keypoints. One way to use image features is through the confidence predictions, similar to previous works \cite{learnable-triangulation, ng-ransac, openpose}.

Finally, we assume that the intrinsic camera parameters and the scale are known.

\section*{E. Ethical Considerations}
\label{appendix-4}

For all of our experiments, we use two well-known, public datasets --- Human3.6M and CMU Panoptic Studio. From the information obtained from the corresponding websites, it is unclear whether the datasets have the IRB approvals. We verified with the authors of the Panoptic Studio that the dataset has the approval. We also contacted the authors of Human3.6M, but did not get the confirmation at the moment of writing.

{\small
\bibliographystyle{ieee}
\bibliography{main}

\begin{thebibliography}{10}\itemsep=-1pt

\bibitem{a-review-of-body-measurement}
K.~Bartol, D.~Bojanić, T.~Petković, and T.~Pribanić.
\newblock A review of body measurement using 3d scanning.
\newblock {\em IEEE Access}, 9:67281--67301, 2021.

\bibitem{3d-pictorial-structures}
V.~Belagiannis, S.~Amin, M.~Andriluka, B.~Schiele, N.~Navab, and S.~Ilic.
\newblock 3d pictorial structures for multiple human pose estimation.
\newblock In {\em 2014 IEEE Conference on Computer Vision and Pattern
  Recognition}, pages 1669--1676, 2014.

\bibitem{self-supervised-with-multiple-view}
A.~Bouazizi, J.~Wiederer, U.~Kressel, and V.~Belagiannis.
\newblock Self-supervised 3d human pose estimation with multiple-view geometry.
\newblock {\em ArXiv}, abs/2108.07777, 2021.

\bibitem{dsac}
E.~Brachmann, A.~Krull, S.~Nowozin, J.~Shotton, F.~Michel, S.~Gumhold, and
  C.~Rother.
\newblock Dsac — differentiable ransac for camera localization.
\newblock {\em 2017 IEEE Conference on Computer Vision and Pattern Recognition
  (CVPR)}, pages 2492--2500, 2017.

\bibitem{less-is-more}
E.~Brachmann and C.~Rother.
\newblock Learning less is more - 6d camera localization via 3d surface
  regression.
\newblock {\em 2018 IEEE/CVF Conference on Computer Vision and Pattern
  Recognition}, pages 4654--4662, 2018.

\bibitem{ng-ransac}
E.~Brachmann and C.~Rother.
\newblock Neural-guided ransac: Learning where to sample model hypotheses.
\newblock {\em 2019 IEEE/CVF International Conference on Computer Vision
  (ICCV)}, pages 4321--4330, 2019.

\bibitem{real-time-3d-pose-smart-edge-sensors}
S.~Bultmann and S.~Behnke.
\newblock Real-time multi-view 3d human pose estimation using semantic feedback
  to smart edge sensors.
\newblock In D.~A. Shell, M.~Toussaint, and M.~A. Hsieh, editors, {\em
  Robotics: Science and Systems XVII, Virtual Event, July 12-16, 2021}, 2021.

\bibitem{openpose}
Z.~Cao, T.~Simon, S.-E. Wei, and Y.~Sheikh.
\newblock Realtime multi-person 2d pose estimation using part affinity fields.
\newblock {\em 2017 IEEE Conference on Computer Vision and Pattern Recognition
  (CVPR)}, pages 1302--1310, 2017.

\bibitem{can-3d-pose-be-learned-from-2d}
D.~Drover, M.~Rohith, C.-H. Chen, A.~Agrawal, A.~Tyagi, and C.~P. Huynh.
\newblock Can 3d pose be learned from 2d projections alone?
\newblock In {\em ECCV Workshops}, 2018.

\bibitem{ransac}
M.~A. Fischler and R.~C. Bolles.
\newblock Random sample consensus: A paradigm for model fitting with
  applications to image analysis and automated cartography.
\newblock {\em Commun. ACM}, 24(6):381–395, June 1981.

\bibitem{mvs-tutorial}
Y.~Furukawa and C.~Hernández.
\newblock {\em Multi-View Stereo: A Tutorial}.
\newblock 2015.

\bibitem{zisserman}
R.~Hartley and A.~Zisserman.
\newblock {\em Multiple View Geometry in Computer Vision}.
\newblock Cambridge University Press, USA, 2 edition, 2003.

\bibitem{epipolar-transformers}
Y.~He, R.~Yan, K.~Fragkiadaki, and S.-I. Yu.
\newblock Epipolar transformers.
\newblock {\em 2020 IEEE/CVF Conference on Computer Vision and Pattern
  Recognition (CVPR)}, pages 7776--7785, 2020.

\bibitem{h36m}
C.~Ionescu, D.~Papava, V.~Olaru, and C.~Sminchisescu.
\newblock Human3.6m: Large scale datasets and predictive methods for 3d human
  sensing in natural environments.
\newblock {\em IEEE Transactions on Pattern Analysis and Machine Intelligence},
  36(7):1325--1339, jul 2014.

\bibitem{learnable-triangulation}
K.~Iskakov, E.~Burkov, V.~Lempitsky, and Y.~Malkov.
\newblock Learnable triangulation of human pose.
\newblock {\em 2019 IEEE/CVF International Conference on Computer Vision
  (ICCV)}, pages 7717--7726, 2019.

\bibitem{gumbel-softmax}
E.~Jang, S.~Gu, and B.~Poole.
\newblock Categorical reparameterization with gumbel-softmax.
\newblock {\em ArXiv}, abs/1611.01144, 2017.

\bibitem{cmu-panoptic}
H.~Joo, T.~Simon, X.~Li, H.~Liu, L.~Tan, L.~Gui, S.~Banerjee, T.~S. Godisart,
  B.~Nabbe, I.~Matthews, T.~Kanade, S.~Nobuhara, and Y.~Sheikh.
\newblock Panoptic studio: A massively multiview system for social interaction
  capture.
\newblock {\em IEEE Transactions on Pattern Analysis and Machine Intelligence},
  2017.

\bibitem{generalizable-approach}
A.~Kadkhodamohammadi and N.~Padoy.
\newblock A generalizable approach for multi-view 3d human pose regression.
\newblock {\em Mach. Vis. Appl.}, 32:6, 2021.

\bibitem{self-supervised-3d-pose-using-multi-view-geometry}
M.~Kocabas, S.~Karagoz, and E.~Akbas.
\newblock Self-supervised learning of 3d human pose using multi-view geometry.
\newblock {\em 2019 IEEE/CVF Conference on Computer Vision and Pattern
  Recognition (CVPR)}, pages 1077--1086, 2019.

\bibitem{8-point}
H.~Longuet-Higgins.
\newblock A computer algorithm for reconstructing a scene from two projections.
\newblock In M.~A. Fischler and O.~Firschein, editors, {\em Readings in
  Computer Vision}, pages 61--62. Morgan Kaufmann, San Francisco (CA), 1987.

\bibitem{sift}
D.~G. Lowe.
\newblock Distinctive image features from scale-invariant keypoints.
\newblock {\em Int. J. Comput. Vision}, 60(2):91–110, Nov. 2004.

\bibitem{concrete-distribution}
C.~J. Maddison, A.~Mnih, and Y.~Teh.
\newblock The concrete distribution: A continuous relaxation of discrete random
  variables.
\newblock {\em ArXiv}, abs/1611.00712, 2017.

\bibitem{a-star-sampling}
C.~J. Maddison, D.~Tarlow, and T.~Minka.
\newblock A* sampling.
\newblock In {\em NIPS}, 2014.

\bibitem{interhand-2.6m}
G.~Moon, S.-I. Yu, H.~Wen, T.~Shiratori, and K.~M. Lee.
\newblock Interhand2.6m: A dataset and baseline for 3d interacting hand pose
  estimation from a single rgb image.
\newblock {\em ArXiv}, abs/2008.09309, 2020.

\bibitem{cross-view-fusion}
H.~Qiu, C.~Wang, J.~Wang, N.~Wang, and W.~Zeng.
\newblock Cross view fusion for 3d human pose estimation.
\newblock {\em 2019 IEEE/CVF International Conference on Computer Vision
  (ICCV)}, pages 4341--4350, 2019.

\bibitem{lightweight-multi-view}
E.~Remelli, S.~Han, S.~Honari, P.~Fua, and R.~Y. Wang.
\newblock Lightweight multi-view 3d pose estimation through camera-disentangled
  representation.
\newblock {\em 2020 IEEE/CVF Conference on Computer Vision and Pattern
  Recognition (CVPR)}, pages 6039--6048, 2020.

\bibitem{stochastic-computation-graphs}
J.~Schulman, N.~Heess, T.~Weber, and P.~Abbeel.
\newblock Gradient estimation using stochastic computation graphs.
\newblock In {\em NIPS}, 2015.

\bibitem{colmap}
J.~L. Schönberger and J.-M. Frahm.
\newblock Structure-from-motion revisited.
\newblock In {\em 2016 IEEE Conference on Computer Vision and Pattern
  Recognition (CVPR)}, pages 4104--4113, 2016.

\bibitem{adaptively-multi-view-transformer}
H.~Shuai, L.~Wu, and Q.~Liu.
\newblock Adaptively multi-view and temporal fusing transformer for 3d human
  pose estimation.
\newblock {\em ArXiv}, abs/2110.05092, 2021.

\bibitem{multiview-bootstrapping}
T.~Simon, H.~Joo, I.~Matthews, and Y.~Sheikh.
\newblock Hand keypoint detection in single images using multiview
  bootstrapping.
\newblock {\em 2017 IEEE Conference on Computer Vision and Pattern Recognition
  (CVPR)}, pages 4645--4653, 2017.

\bibitem{view-invariant-probabilistic-embedding}
J.~J. Sun, J.~Zhao, L.-C. Chen, F.~Schroff, H.~Adam, and T.~Liu.
\newblock View-invariant probabilistic embedding for human pose.
\newblock In A.~Vedaldi, H.~Bischof, T.~Brox, and J.-M. Frahm, editors, {\em
  Computer Vision -- ECCV 2020}, pages 53--70, Cham, 2020. Springer
  International Publishing.

\bibitem{rethinking-pose-in-3d}
D.~Tom{\`e}, M.~Toso, L.~Agapito, and C.~Russell.
\newblock Rethinking pose in 3d: Multi-stage refinement and recovery for
  markerless motion capture.
\newblock {\em 2018 International Conference on 3D Vision (3DV)}, pages
  474--483, 2018.

\bibitem{total-capture}
M.~Trumble, A.~Gilbert, C.~Malleson, A.~Hilton, and J.~Collomosse.
\newblock Total capture: 3d human pose estimation fusing video and inertial
  sensors.
\newblock In {\em 2017 British Machine Vision Conference (BMVC)}, 2017.

\bibitem{voxelpose}
H.~Tu, C.~Wang, and W.~Zeng.
\newblock Voxelpose: Towards multi-camera 3d human pose estimation in wild
  environment.
\newblock In {\em European Conference on Computer Vision (ECCV)}, 2020.

\bibitem{cpm}
S.-E. Wei, V.~Ramakrishna, T.~Kanade, and Y.~Sheikh.
\newblock Convolutional pose machines.
\newblock {\em 2016 IEEE Conference on Computer Vision and Pattern Recognition
  (CVPR)}, pages 4724--4732, 2016.

\bibitem{monocular-total-capture}
D.~Xiang, H.~Joo, and Y.~Sheikh.
\newblock Monocular total capture: Posing face, body, and hands in the wild.
\newblock {\em 2019 IEEE/CVF Conference on Computer Vision and Pattern
  Recognition (CVPR)}, pages 10957--10966, 2019.

\bibitem{simple-baselines}
B.~Xiao, H.~Wu, and Y.~Wei.
\newblock Simple baselines for human pose estimation and tracking.
\newblock In {\em ECCV}, 2018.

\bibitem{deep-kinematics-analysis}
J.~Xu, Z.~Yu, B.~Ni, J.~Yang, X.~Yang, and W.~Zhang.
\newblock Deep kinematics analysis for monocular 3d human pose estimation.
\newblock In {\em Proceedings of the IEEE/CVF Conference on Computer Vision and
  Pattern Recognition (CVPR)}, June 2020.

\bibitem{learning-to-find-good-correspondences}
K.~M. Yi, E.~Trulls, Y.~Ono, V.~Lepetit, M.~Salzmann, and P.~Fua.
\newblock Learning to find good correspondences.
\newblock {\em 2018 IEEE/CVF Conference on Computer Vision and Pattern
  Recognition}, pages 2666--2674, 2018.

\end{thebibliography}
}

\end{document}